\documentclass[journal]{IEEEtran}
\usepackage{amsmath}
\usepackage{amssymb,amsmath,amsthm}
\usepackage{graphicx}
\usepackage{amssymb}
\usepackage{cite}
\usepackage{cite}
\usepackage{xfrac}
\usepackage{url}
\usepackage{bm}
\usepackage{url}
\usepackage{float}
\usepackage{lipsum}

\newtheorem{thm}{Theorem}
\newtheorem{lem}{Lemma}
\newtheorem{prop}{Proposition}

\newtheorem{defin}{Definition}
\newenvironment{example}[2][Example]{\begin{trivlist}
\item[\hskip \labelsep {\bfseries #1}\hskip \labelsep {\bfseries #2}]}{\end{trivlist}}
\newenvironment{rem}[1][Remark]{\begin{trivlist}
\item[\hskip \labelsep {\bfseries #1}]}{\end{trivlist}}
\newcommand{\bd}{\begin{defin}}
\newcommand{\ed}{\end{defin}}
\newcommand{\bl}{\begin{lem}}
\newcommand{\el}{\end{lem}}
\newcommand{\br}{\begin{rem}}
\newcommand{\er}{\end{rem}}
\newcommand{\bt}{\begin{thm}}
\newcommand{\et}{\end{thm}}
\newcommand{\be}{\begin{example}}
\newcommand{\ee}{\end{example}}
\newcommand{\bp}{\begin{prop}}
\newcommand{\ep}{\end{prop}}

\newcommand{\bqn}{\begin{eqnarray}}
\newcommand{\eqn}{\end{eqnarray}}

\newcommand{\inn}{\,\in\,}

\def\inn{\,\in\,}

\def\1{\hbox{\rm\rlap {1}\hskip.03in{\rom I}}}
\def\Bbbone{{\rm1\mathchoice{\kern-0.25em}{\kern-0.25em}
{\kern-0.2em}{\kern-0.2em}I}}

\ifCLASSINFOpdf
\else
\fi
\begin{document}
%

%
%
%

\title{Effective Connectivity-Based Neural Decoding: A Causal Interaction-Driven Approach }
\author{Saba~Emrani
        and~Hamid~Krim      
\thanks{S. Emrani and H. Krim are with the Department
of Electrical and Computer Engineering, North Carolina State University, Raleigh,
NC, 27606 USA e-mails: semrani@ncsu.edu, ahk@ncsu.edu.}
\thanks{This work was supported by National Science Foundation EEC-1160483.}}

\maketitle

\begin{abstract}
We propose a geometric model-free causality measure
based on multivariate delay embedding that can efficiently detect linear and nonlinear causal interactions between time series with no prior information.
We then exploit the proposed causal interaction measure in real MEG data analysis. The results are used to construct effective connectivity maps of brain activity to decode different categories of visual stimuli. 
Moreover, we discovered that the MEG-based effective connectivity maps as a response to structured images exhibit more geometric patterns, as disclosed by analyzing the evolution of toplogical structures of the underlying networks using persistent homology. Extensive simulation and experimental result have been carried out to substantiate the capabilities of the
proposed approach. 
\end{abstract}


%
\IEEEpeerreviewmaketitle

\section{Introduction}
%
%
%
%

\IEEEPARstart{M}{} agnetoencephalography (MEG)  and Electroencephalography (EEG) are the most common noninvasive neurophysiological techniques for direct measurement of brain function. Neural activities in the brain generate electric and magnetic fields, that are detected by EEG and MEG measurements with a high temporal resolution. EEG sensors measure the aggregate of the simultaneous activity of millions of neurons on the same spatial orientation in the brain. Sensitive MEG sensors detect the weak magnetic fields outside the skull caused by synchronized neuronal currents. In addition to high temporal resolution,  MEG provides excellent spatial resolution by measuring whole-head neural signals using 100 to more than 300 sensors, making it a preferred tool to study brain connectivity \cite{baillet2001electromagnetic}. 
%
%
%
The main objective of MEG data analysis is to derive spatio-temporal structures or deduce connectivity patterns between various brain regions from extensive and intricate time series. MEG signal processing has many applications in clinical and experimental research settings, particularly in perceptual and cognitive brain processes \cite{grasman2004frequency, gomez2009disturbed}. However, the analysis of MEG data is  exceptionally challenging due to its complex characteristics including high dimensionality, 
nonstationarity across time, trials, conditions and subjects, low signal to noise ratio,  nonlinearity of features with respect to sensor channels, as well as simultaneous presence of other patterns in the data such as event related potentials/fields and oscillations \cite{Wu2016}.

Traditional signal processing techniques for MEG/EEG analysis
are predominantly focused on digital filtering, spectral analysis and source
separation \cite{matani2003hierarchical, signoretto2012classification, hornero2008spectral, vigario2000independent}. MEG-based decoding of the brain states has however attracted great attention over the past decade.
Several machine learning techniques have been used for MEG decoding including linear discriminant analysis (LDA) \cite{waldert2008hand, pfurtscheller2000current, shenoy2008generalized}, support vector machine \cite{montazeri2008meg, shenoy2008generalized} and common spatial pattern (CSP) \cite{lu2010regularized}. The objective of the majority of decoding methods is identification of optimal discriminant patterns that differentiate distinct brain states with high accuracy. These patterns can be obtained by analysis of signals collected by extensively scattered sensors on the brain.

In this study, we focus on a connectivty-based decoding of different visual stimuli using MEG recordings. Brain networks comprises spatially distributed but functionally connected regions which process information. Several measures have been defined for the quantification of neural interactions to result in three main categories of connectivity: 1) Anatomical or structural connectivity that shapes the \textit{connectome} by analyzing patterns of synaptic contacts between adjacent neurons or fiber tracks between neuronal populations in spatially distant brain areas. 2) Functional connectivity defined as the
temporal correlations or statistical dependencies of neuronal activity patterns between distinct and usually remotely separated regions. 3) Effective connectivity which is of interest in this paper, describes directional activity profiles of one neural system over another, hence representing causal interactions between activated brain
regions. Effective connectivity may be viewed as a combination of functional connectivity and intrinsic structural model of a network with causal effects. Since causes precede effects in time, causal effects can be inferred from time series analysis. Causal interaction between two brain regions is deduced if a significant potentially time-lagged, influence between the corresponding time series is discovered.

Various connectivity measures for quantifying the causal influence and for evaluating effective connectivity have been proposed. Many of these have important limitations.
These approaches are mainly formulated using autoregressive (AR) models such as Granger causality \cite{granger1969investigating}, the cross spectrum such as coherence and phase slope index \cite{nolte2008robustly}, or directed transfer function \cite{kaminski1991new}, and partial directed coherence methods \cite{baccala2001partial, haufe2009sparse}.
Most of these measures are rooted in linear regression modeling and cannot identify the nonlinear interactions between MEG/EEG signals. For example Granger causality as a widely used tool in many valuable applications in neuroscience \cite{brovelli2004beta, he2011electrophysiological} is based on a linear bivariate autoregressive model. It has also been recently extended based on a nonlinear autoregressive model by
fitting a nonlinear polynomial model \cite{zhao2013new}.
This however requires prior knowledge of the system model.
Since the interacting neural structures are unknown in advance, a model-free causal interaction measure which accounts for nonlinear as well as linear causal interactions is highly desirablel.

In this paper, we present a geometric, non-parametric and model-free causality measure based on manifold learning. This can efficiently detect linear and nonlinear causal interactions between time series. This framework is based on the concept of multivariate delay embedding. The mathematical foundation of delay coordinate embedding method was first proposed by Takens \cite{Takens} to capture the dynamics of
a scalar time series in a higher dimensional
space. There is a large body of literature on the application
of  delay embedding to dynamical systems with chaotic attractors \cite{Abar,Kantz}.
We have used univariate time delay embedding as an efficient tool for identification of harmonic patterns in signals as well as estimating their spectral characteristics \cite{meSPL14},\cite{meEUSIPCO}. In this study, we extend the delay embedding framework to more than one modality data by developing an approach for evaluating causal interactions using multivariate delay embedding. Our proposed causality measure is based on the fractal dimension of the point cloud constructed by multivariate delay embedding. 
We subsequently use our causal interaction measure to analyze MEG time series recorded as a brain's response to visual stimuli, to construct as a result effective connectivity maps and use these maps to decode different categories of visual stimuli. There has been no study to date on effective connectivity based decoding of visual stimulus using MEG data.
Furthermore, we discover that the MEG data based effective connectivity maps as a response to structured images exhibit more geometric patterns, as revealed by analyzing the evolution of toplogical patterns of the underlying networks using persistent homology.

The remainder of the paper is organized as follows. In Section \ref{Pframe}, we describe in detail our proposed analysis framework. The multivarie delay embedding approach will first be described and formulated in Section \ref{MVDE}. Fractal-based dimension estimation methods are presented in Section \ref{DimE}, including Box-Counting technique which we use to validate our causality measure and correlation dimension estimation fully exploited in multivariate delay embedding point clouds for uncovering their dimensions. In section \ref{CIM}, we present our causal interaction measure based on fractal dimension of multivariate delay embeddings. We will subsequently explain in Section \ref{ConMaps}, how we construct comprehensive effective connectivity maps by exploiting our proposed causality measure for MEG data measurements.
In Section \ref{CIMResults}, we validate the introduced causal interaction measure using simulated data as well as experimental MEG data. Several simulation examples are provided to evaluate the capability of the proposed measure in detecting directed, linear and nonlinear causal interactions in both noise free and noisy environment. Moreover, analysis of causal interactions in recorded MEG data recorded during during a brain's active and rest mode. In Section \ref{CIMResults}, we decode three categories of images of artificial objects, natural sceneries and soccer game using causal interactions of MEG time series and show promising results compared to the existing methods. We then use persistent homology in Section \ref{PHEC} for analysis of effective connectivity maps by evaluation of their topological features over varying scales and show the existence of geometric structures in the connectivity maps corresponding to more structured visual stimuli. Finally, Section \ref{Conclu} will conclude the paper and discuss the future research directions.



%


\section{Causality by Delay Embeddings}\label{Pframe}
\subsection{Multivariate Delay Embedding}\label{MVDE}

Univariate delay embedding for a time series $\{x_n \}_{n=1}^N$ 
is defined as $\{X_n \}=(x_n,x_{n-\tau},\dotsc ,x_{n-(m-1)\tau} )$, where $\tau$ is the time delay, and $m$ is the embedding dimension. With a similar intuition, multivariate delay embedding of $p$ time series $\{x_{i,n} \}_{n=1}^N, i=1,2,\dotsc,p$ can be defined as 
\begin{multline}
\{X_n\}=(x_{1,n},x_{1,n-\tau},\dotsc,x_{1,n-(m_1-1)\tau},x_{2,n},x_{2,n-\tau},\dotsc,\\
x_{2,n-(m_2-1)\tau},\dotsc,x_{p,n-(m_p-1)\tau}),
\end{multline}
where $\bm{m}=(m_1,m_2,\dotsc,m_p)$ is the embedding dimension vector, and determines the number of components included from each time series. The ambient dimension of the embedding space is therefore $M= \sum_{i=1}^{p} m_i$. In a more general case, different $\tau$'s for each time series can be used as
\begin{multline}
\{X_n\}=(x_{1,n},x_{1,n-\tau_1},\dotsc,x_{1,n-(m_1-1)\tau_1}, x_{2,n},x_{2,n-\tau_2},\dotsc,\\
x_{2,n-(m_2-1)\tau_2},\dotsc,x_{p,n-(m_p-1)\tau_p}),
\end{multline}
where $\tau=\left(\tau_1, \tau_2,\dotsc,\tau_p \right)$ is the time delay vector for $p$ time series. Non-uniform multivariate delay embedding which is the most general embedding technique can be defined for $m_i$ varying delays denoted by $l_{ij},j=1,\dotsc,m_i$ at each time series $\{x_{i,n} \}_{n=1}^N$ as follows,
\begin{multline}
\{X_n\}=(x_{1,n-l_{11}},x_{1,n-l_{12}},\dotsc,x_{1,n-l_{1m_1}},x_{2,n-l_{21}},\dotsc, \\ x_{p,n-l_pm_p}).
\end{multline}
\subsection{Fractal-based Dimension Estimation}\label{DimE}
While the multivariate embedding point cloud is embedded in space $\mathbb{R}^M$, its intrinsic dimension is not necessarily $M$. The intrinsic dimension is defined as the minimum number of free variables needed to represent the data (degrees of freedom) with no loss of information. Equivalently, the intrinsic dimension of a dataset in $\mathbb{R}^M$ is equal to $d$ if its elements lie totally within a $d$-dimensional subspace of $\mathbb{R}^M$, where $d < M$ \cite{krishnaiah1982}.

There are two main approaches for estimating intrinsic dimension of a dataset. Local techniques estimate the intrinsic dimension using the information in the local neighborhood of data points, while global approaches use the whole dataset and unfold it in a $d$-dimensional space\cite{jain1988}. Fractal-based techniques are global methods that can provide a non-integer value as the intrinsic dimension of data. A variety of techniques have been previously proposed for fractal dimension estimation \cite{klinkenberg94}. The Box-Counting method is the most common definitions for fractal dimension. Box-counting dimension $D_B$ of a set $\bm{S}$ in $\mathbb{R}^M$ is defined as follows \cite{belussi98}:

\textit{Fact:} If $\nu (r)$ is the number of hyper-boxes of size $r$ required to cover $\bm{S}$, then the box-counting fractal dimension of $\bm{S}$ is defined as 
\begin{equation}\label{Box}
D_B = \lim _{r \to 0} \frac{\ln(\nu(r))}{\ln(1/r)}.
\end{equation}

Suppose the $M$-dimensional dataset has an intrinsic dimension of $d$ and there are adequately large number of data points. As $r$ decreases, the relationship between the number of hyperboxes $\nu(r)$ needed to cover $\bm{S}$ is proportional to $1/r^d$. Hence, we have 
\begin{equation}\label{dim}
\nu(r) \propto 1/r^d \quad \textit{or}  \quad  \nu(r) = c/r^d ,
\end{equation}
where $c$ is a constant. Applying $\ln$ to both sides of Equation (\ref{dim}) yields
\begin{equation}
\ln(\nu(r)) = \ln c +d \ln(\frac{1}{r}).
\end{equation}
Therefore, the intrinsic dimension $d$ can be obtained as
\begin{equation}
d= \frac{\ln (\nu(r))}{\ln(\frac{1}{r})} - \frac{\ln c}{\ln(\frac{1}{r})}.
\end{equation}
The second term $(\ln c / \ln(\frac{1}{r}))$ vanishes as $r$ approaches zero. Thus, the box-counting fractal dimension of the data provides an appropriate estimation of its intrinsic dimension. 
Practically, datasets include a finite number of points. Therefore, the increase in $\nu(r)$ decelerates as $r$ decreases and reaches a state of little or no change after a time. Therefore, the dimension is calculated using the plot of $\nu(r)$ with respect to $1/r$ and computing the slope of its linear part.

Although many algorithms have been proposed for box-counting dimension estimation, their complexity increases exponentially with the dimension of the dataset. Alternatively, correlation dimension \cite{grassberger2004} is a computationally simple approach for estimating intrinsic dimension. 

\textit{Fact:} Suppose $\bm{S} = \{s_1, s_2,....s_n\}$ is a set of data points in $\mathbb{R}^M$. The correlation integral $C_n(r)$ is defined as 
\begin{equation}
C(n,r)= \frac{2}{n(n-1)}\sum_{i=1}^n \sum_{j=i+1}^n I(\parallel s_j -s_i \parallel \le r),
\end{equation}
where $I(.)$ is an indicator function, i.e. $I(\lambda) =1$ if and only if condition $\lambda$ holds and zero otherwise. Also, $\parallel s_j -s_i \parallel $ denotes the Euclidean distance between data points $s_j$ and $s_i$. $C(n,r)$ is basically the probability of a pair of points having a distance smaller than or equal to $r$. 
The correlation fractal dimension $D_c$ of $\bm{S}$ is then defined as 
\begin{equation}
D_c = \lim _{n \to \infty} \lim _{r \to 0} \frac{\ln (C(n,r))}{\ln r}.
\end{equation}
Similar to box-counting dimension, the correlation dimension is also estimated in practice by using the plot of $\ln C(n,r)$ with respect to $r$ and by computing the slope of its linear part. Due to the noisy nature of real data, the difference between the correlation dimension and box-counting dimension is negligible as proved in \cite{camastra2003}.
\subsection{CIM:Causal Interaction Measure}\label{CIM}

Consider three time series $X=\{x_n \}_{n=1}^N$, $Y=\{y_n \}_{n=1}^N$ and $Z=\{z_n \}_{n=1}^N$ and their multivariate delay embedding vectors $\{x_n, y_{n-\tau_1} \}$ and $\{x_n, z_{n-\tau_2} \}$.

\begin{thm}\label{thm1}
The causal interaction between time series $X=\{x_i \}_{i=1}^N$ and  $Y=\{y_i \}_{i=1}^N$ with delay $\tau_1$ is higher than the causal interaction between time series $X=\{x_i \}_{i=1}^N$ and  $Z=\{z_i \}_{i=1}^N$ with delay $\tau_2$  if and only if the dimension of the point cloud represented by $(x_n, y_{n-\tau_1} )$ is lower than the dimension of the point cloud $(x_n, z_{n-\tau_2} )$.

\end{thm}
The proof of Theorem  \ref{thm1} can be found in the Appendix. In order to find the delay corresponding to the strongest causal interaction between time series, we follow a similar analogy. If we simply replace $Z$ with $Y$ in Theorem \ref{thm1}, we can conclude the following: The causal interaction between time series $X=\{x_i \}_{i=1}^N$ and  $Y=\{y_i \}_{i=1}^N$ with delay $\tau_1$ is higher than the causal interaction between them with delay $\tau_2$  if and only if the dimension of the point cloud represented by $\{x_n, y_{n-\tau_1} \}$ is lower than the dimension of the point cloud $\{x_n, y_{n-\tau_2} \}$. 
 
We denote the dimension of the delay embedding point cloud $\{x_n, y_{n-\tau_1} \}$ as $d_{XY}$ and propose $1/d_{XY}$ as a causal interaction measure (CIM) between time series $X=\{x_i \}_{i=1}^N$ and  $Y=\{y_i \}_{i=1}^N$ with delay $\tau_1$. In other words, there is a directed information flow from $X$ to $Y$ with delay $\tau_1$ with strength $1/d_{XY}$. We use the box-counting definition of fractal dimension to prove the Theorem and exploit the correlation dimension estimation method to compute the dimensions of the point clouds (Section \ref{DimE}). The calculated fractal dimension and the final CIM values are therefore not necessarily integer numbers. 
%
For example, when we construct two-dimensional multivariate emebedding, the smallest dimension value corresponds to a curve and occurs when the highest interaction exists between two time series. Also, the largest possible dimension equals the dimension of the ambient space  which corresponds to the case where there is no interaction between the two time series (2 in this case).

\subsection{Efficient Estimation: A Computational Approach }

As investigating all possible combinations of the whole time series is computationally complex, we use a progressive geometrical method similar to the one suggested in  \cite{vlachos2010nonuniform} to build the embedding vectors. First, we determine a maximum lag according to the existing practical estimations of maximum delays in the causal interaction between time series $\{x_{i,n} \}_{n=1}^N, i=1,2,\dotsc,p$ as $L_i,i=1,…,p$. We will then construct a candidate embedding vector using all $L_1+L_2+ \dotsc +L_p$ elements as 
\begin{equation}
\bm{B}=\{x_{1,n},x_{1,n-1},\dotsc,x_{1,n-L_1},x_{2,n},x_{2,n-1},\dotsc,x_{p,n-L_p}\}.
\end{equation}
The embedding vector $\bm{b}$ to be chosen is a subset of B including the time series with causal interaction with the corresponding delays. Similar to the concept of Granger causality, we can infer that the selected embedding vector is to increase our knowledge about the future of the system in one or several steps ahead. Suppose that for time series $\{x_{i,n} \}_{n=1}^N$, $T_i$ future steps need to be investigated. The future state of the system is then denoted by 
\begin{equation}
X_F=\left(x_{i,n+1},x_{i,n+2},…,x_{i,n+T_1} \right).
\end{equation}
One possible solution for constructing the embedding vector $\bm{b}$ starts with an empty vector $\bm{b_0}$. Suppose that the selected embedding vector at step $(j-1)$ is 
$\bm{b_{j-1}}=(x_1^\star,x_2^\star,…,x_{j-1}^\star)$. At step $j$, the element $x_j^\star \in \bm{B} \setminus \bm{b_{j-1}}$ will be added to $\bm{b_{j-1}}$ to build a candidate vector for the next step $\bm{b_{j}}$. We will then analyze the point cloud of the delay embedding vector $\bm{b_{j}}$ with $x_1^\star,x_2^\star,…,x_{j-1}^\star$ and $x_j^\star$ as the corresponding coordinates. If the point cloud forms a manifold with an intrinsic dimension less than the ambient space, we conclude that there is interaction between the components of $\bm{b_{j}}$. As a result, we include $x_j^\star$ in the embedding vector. Otherwise, $x_j^\star$ will be skipped and the next element will be taken into consideration.

This reconstruction scheme is based on Theorem \ref{thm1}, meaning if there is any interaction between time series, the intrinsic dimension of the their corresponding multivariate embedding is less than the dimension of the ambient space.  The estimation of the dimensionality of the delay-coordinate embedding point clouds constructed using real data sets, will be performed using correlation dimension estimation method. We will then need a threshold on the obtained fractal dimension that can be experimentally established. The components that establish a dimension less than the selected threshold will be included in the embedding vector.

The approach described above is a general formulation and can describe a variety of models. Our previous framework of univariate delay embedding \cite{meSPL14, meEUSIPCO} can be obtained when $X_F$ and $\bm{B}$ include elements from the same time series. Cross modeling happens when $X_F$ has elements from one time series and $\bm{B}$ from another one. The more general case is mixed modeling when $X_F$ has elements from one time series and $\bm{B}$ from all the time series. The most comprehensive condition is full modeling when $X_F$ and $\bm{B}$ both have elements from all time series. In this paper, we mainly use cross modeling for MEG recordings to construct effective connectivity maps.

\subsection{Effective Connectivity Maps}\label{ConMaps}

In this Section, we describe our exploitation of the proposed causal interaction measure (CIM) in MEG time series data analysis. The result is destined to construct a comprehensive map of effective connectivities of brain activity. The connectivity maps include information about the spatial location of all the sensors as well as their causal connectivity patterns.
Since the brain response time to visual stimuli is extremely small, we create connectivity maps in small time windows taken from longer MEG recordings, resulting in time varying connectivity patterns. After windowing the time series, for each sensor measurement $\{x_{k,n}\}_{n=1}^N$, we construct its pairwise two dimensional multivariate delay embedding with each of the other sensors $j \in \{1,2,...N_s\}\setminus k$ as $(x_{k,n}, x_{j,n-\tau_i})$, where $N_s$ is the number of MEG sensors. For each $j$, we use different delays selected from a set of possible lags $\tau_i \in T$, where the maximum lag is determined by the maximum possible causality delay between brain regions. We then calculate the dimension of all the constructed point clouds for each pair of sensors $(k,j)$ and select the point cloud with the smallest dimension $(x_{k,n}, x_{j,n-\tau_{kj}})$. The dimension of this point cloud is represented by
\begin{equation}\label{weight}
d_{kj} = \displaystyle{\min_{\tau_i \in T} dim[(x_{k,n}, x_{j,n-\tau_i})]}.
\end{equation}
$1/d_{kj}$ represents the strength of causal interaction between sensors $j$ and $k$ with direction from sensor $j$ to sensor $k$ and $\tau_{kj}$ is the corresponding delay. We form the connectivity maps as a weighted directed graph, where nodes are the sensors distributed on the brain with their accurate spatial locations and $1/d_{kj}$ is the weight of the edge from node $j$ to node $k$. The adjacency matrix of the effective connectivity maps can be defined as $\bm{A}=[a_{kj}]$, where 

\begin{equation}\label{Adj}
  a_{kj} = \left\lbrace  \begin{array}{ccc}
                               1/d_{kj} &  & k \neq j ,   \\
                               0 &  & \text{otherwise,}
                             \end{array}
                             \right.
\end{equation}
As generally $d_{kj} \neq d_{jk}$, $\bm{A}$ is not necessarily symmetric. Note that the original connectivity maps defined above are fully connected graphs with high weights corresponding to strong causal connections and very small weights assigned to the pairs with small or no causal interactions. The matrix of the pairwise lags in the connectivity maps is defined as $L=[l_{kj}]$. 
\begin{equation}\label{Lag}
 l_{kj} = \left\lbrace  \begin{array}{ccc}
                               \tau _{kj} &  & k \neq j ,   \\
                               0 &  & \text{otherwise,}
                             \end{array}
                             \right.
\end{equation}
\section{Causal Interaction Measure: Experimental Validation}\label{CIMResults}
In this section, we illustrate the performance of the causal interaction measure for both simulated data and experimental MEG signals. 

\subsection{Simulated Data}
In order to substantiate the ability of the proposed causal interaction measure to identify linear and nonlinear directed interactions within different delays, we analyze various synthetic data including an example with linear directed flow, an auto-regressive process and unidirectional H\'enon map.

\begin{example}
1\textbf{:} In this example, we test the capability of our proposed method in detecting simple linear directed causal interaction by considering a process fully dependent on another one. Time series $\{y_i\}_{i=1}^{N}$ is causally affected by series $\{x_i\}_{i=1}^{N}$ with delay $1$, while $\{x_i\}_{i=1}^{N}$ is independent, as expressed by the following equation. 
\begin{equation}\label{g}
     \begin{array}{c}
x_i = w_x, \\
y_i =a x_{i-1},
\end{array}                            
\end{equation}
where $w_x$ is zero mean white Gaussian noise with standard deviation of $1$. We generated time series of length 180 samples, with $a=0.5$, constructed multivariate delay embeddings $(x_n, y_{n-1})$ and $(y_n, x_{n-1})$ and calculated the dimensions of the two point clouds for 200 different realizations of the random noise. 
Figure \ref{Ex1_Dim} illustrates the estimated dimensions of the point clouds $(x_n, y_{n-1})$ and $(y_n, x_{n-1})$  in red and blue, respectively. Clearly, the dimension of $(y_n, x_{n-1})$ is much lower than $(x_n, y_{n-1})$, showing strong information flow from $X$ to $Y$ with delay 1. The average dimension over all realizations for $(x_n, y_{n-1})$ and $(y_n, x_{n-1})$ are 1.84 and 0.98, respectively. Therefore, the CIM value equals 0.54 for information flow from $Y$ to $X$ and  1.02 for information flow from $X$ to $Y$, showing a strong flow from $X$ to $Y$ with delay 1. This example validates the ability of the proposed method in detecting the direction of linear causal interaction between two time series. Moreover, the average dimensions of the multivariate delay embeddings for values of delays other than 1 are also much greater than those for $(y_n, x_{n-1})$. Specifically, the average dimension for $(x_n, y_{n-2})$ and $(y_n, x_{n-2})$ are 1.85. This validates the capability of our method in identifying the correct delay in causal interaction.
\begin{figure}[tb] 
	\centering
	\includegraphics[width=1\linewidth]{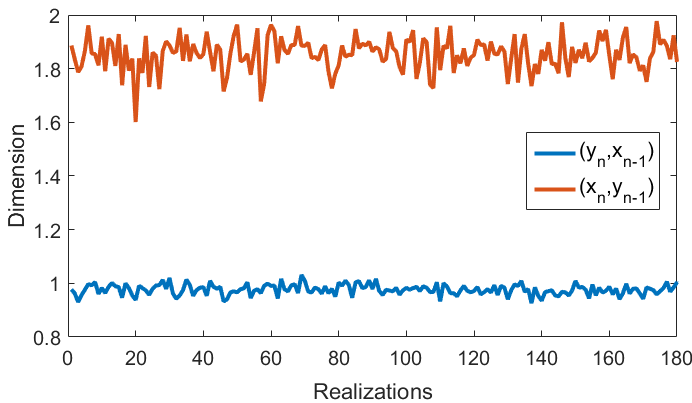}
	\caption{Estimated dimensions of multivariate delay embeddings of simulated data with directed linear causal interaction presented in Example 1. Blue: dimension of $(y_n, x_{n-1})$, red: dimension of $(x_n, y_{n-1})$.
	}
	\label{Ex1_Dim}
\end{figure}

\end{example}

\begin{figure}[tb]
	\centering
	\includegraphics[width=1\linewidth]{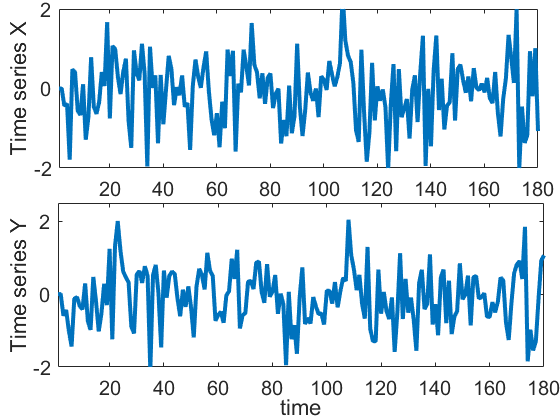}
	\caption{An example of a simulated pair of time series with directed linear causal interactions presented in Equation (\ref{Ex2_Eq}). 
	}
	\label{Ex2_TS}
\end{figure}

\begin{figure}[tb]
	\centering
	\includegraphics[width=1\linewidth]{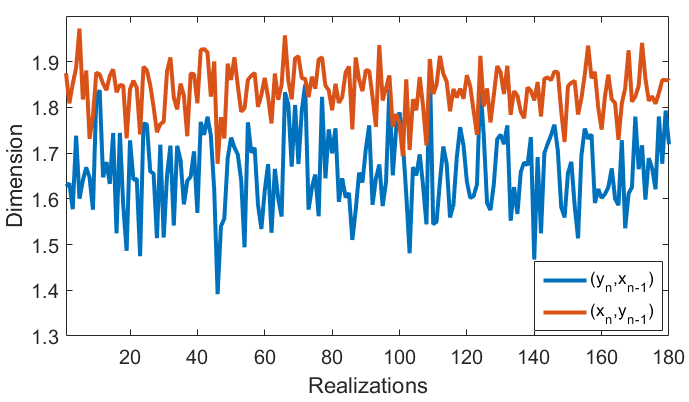}
	\caption{Estimated dimensions of multivariate delay embeddings of the simulated auto-regressive process with directed linear causal interaction presented in Example 2. Blue: dimension of $(y_n, x_{n-1})$, red: dimension of $(x_n, y_{n-1})$
	}
	\label{Ex2_Dims}
\end{figure} 

\begin{example}
2\textbf{:}  In this example, we validate the capability of the proposed method to identify directed linear causal interactions in a first order auto-regressive process, where the second time series is driven by the first one according to the following equation:
\begin{equation}\label{Ex2_Eq}
\begin{array}{c}
                               x_i =0.5 x_{i-1}+u_i,        \\
                               y_i = 0.2 y_{i-1}+0.8 x_{i-1} + v_i,
                             \end{array}   
\end{equation}
where $u_i$ and $v_i$ represent independent white Gaussian noise with standard deviation of $1$ and $0.3$, respectively. Each series consists of 180 samples, and 200 realizations of the auto-regressive process are implemented. One realization of a pair of simulated time series is presented in Figure \ref{Ex2_TS}. 
We constructed two dimensional multivariate delay embeddings of two time series $\{x_i\}_{i=1}^{N}$ and $\{y_i\}_{i=1}^{N}$. The dimension of the multivariate delay embedding $(y_n,x_{n-1})$ is less than dimension of $(x_n,y_{n-1})$ as illustrated in Figure \ref{Ex2_Dims} for different realizations. The average dimension over all realizations for $(x_n, y_{n-1})$ and $(y_n, x_{n-1})$ are 1.83 and 1.65, respectively. Therefore, the CIM value equals 0.54 for information flow from $Y$ to $X$ and 0.61  for information flow from $X$ to $Y$, confirming the existence of a stronger information flow from $\{x_i\}_{i=1}^{N}$ to $\{y_i\}_{i=1}^{N}$ with delay 1, validating the proposed method in identifying directed linear causal interaction. 

\end{example}

 \begin{figure}[tb]
 	\centering
 	\includegraphics[width=1\linewidth]{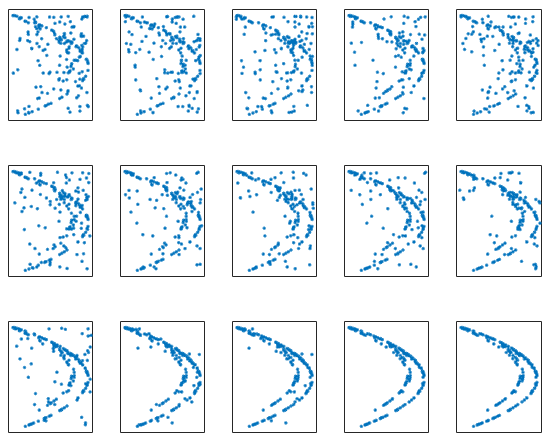}
 	\caption{ H\'enon system's point cloud as the coupling strength increases in the range $[0.01,0.6]$.
 	}
 	\label{Ex3_Henon}
 \end{figure} 
 
 \begin{figure}[tb]
 	\centering
 	\includegraphics[width=0.9\linewidth]{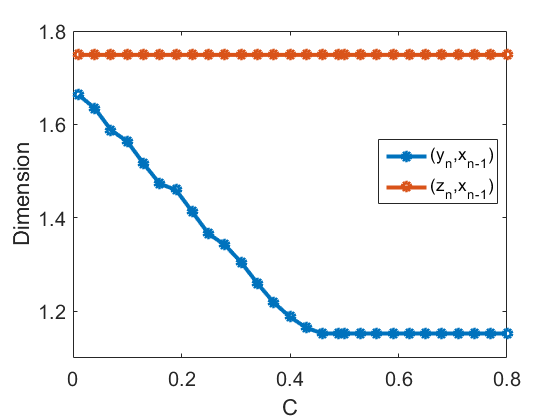}
 	\caption{The dimension of the multivariate delay embedding $(y_n,x_{n-1})$ for coupled H\'enon map for different coupling strengths in $[0.01,0.6]$, compared with that of $(z_n,x_{n-1})$, where $\{z_i\}_{i=1}^N$ and $\{x_i\}_{i=1}^N$ are independent
 	}
 	\label{Ex3_Dims}
 \end{figure} 

\begin{example}
3\textbf{:} In this example, we validate the effectiveness of the proposed method in detecting nonlinear causal interactions and in identify different interaction strengths using coupled H\'enon system with various coupling strengths. The coupled H\'enon map is given by  

\begin{equation}\label{Ex3_Eq}
\begin{array}{c}
                               x_i = 1.4-x^2_{i-1}+0.3x_{i-2}        \\
                               y_i = 1.4-(C y_{i-1}x_{i-1}+(1-C)y^2_{i-1})+0.3 y_{i-1},
                             \end{array}   
\end{equation}
where $C$ is the coupling strength and in this experiment takes values in $[0.01,0.6]$. For each value of $C$, we construct two dimensional delay embeddings for time series of length 200 samples with delay $1$ as $(y_n,x_{n-1})$, shown in Figure \ref{Ex3_Henon}. The dimension of the point cloud for each value of coupling strength $C$ is illustrated in Figure \ref{Ex3_Dims}. Clearly, the dimension of the delay embedding decreases as the driving strength increases. These results demonstrate the capability of our method in determining the strength of causal interactions. Moreover, for comparison we have shown in Figure \ref{Ex3_Dims} the dimension of the point cloud $(z_n,x_{n-1})$ for a nonlinear system given by
\begin{equation*}
z_i = \sin(i) +1.5 \sin(z_{i-1})) +0.6.
\end{equation*}
Since $\{x_i\}_{i=1}^{N}$ and $\{z_i\}_{i=1}^{N}$ are completely independent and there is no interaction between them, the dimension of $(z_n,x_{n-1})$ is higher than all dimensions obtained for H\'enon system as depicted in Figure \ref{Ex3_Dims}. We add white Gaussian observation noise with SNR = 20 dB to each time series and repeat the experiment. In the presence of noise, the multi-variate embedding point clouds change and the dimension of $(y_n, x_{n-1})$ slightly increases but the results in detecting the causal interaction, and its strength are similar to the time series with no noise as shown in Figure \ref{Ex3_DimsN}.

 \begin{figure}[tb]
      \centering
      \includegraphics[width=0.9\linewidth]{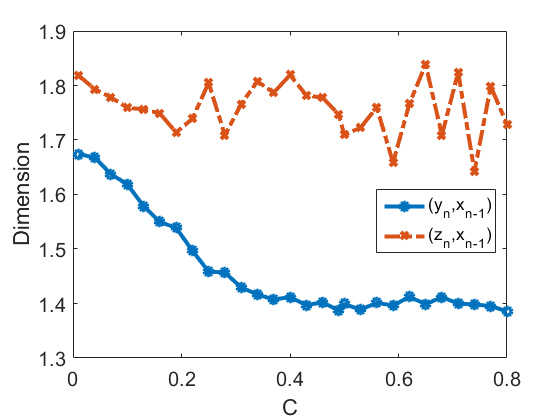}
      \caption{The dimension of the multivariate delay embedding $(y_n,x_{n-1})$ for coupled H\'enon map in the presence of noise for different coupling strengths in $[0.01,0.6]$, compared with that of $(z_n,x_{n-1})$, where $\{z_i\}_{i=1}^N$ and $\{x_i\}_{i=1}^N$ are independent 
      }
       \label{Ex3_DimsN}
   \end{figure} 

\end{example}

\begin{figure}[tb]
	\centering
	\includegraphics[width=0.9\linewidth]{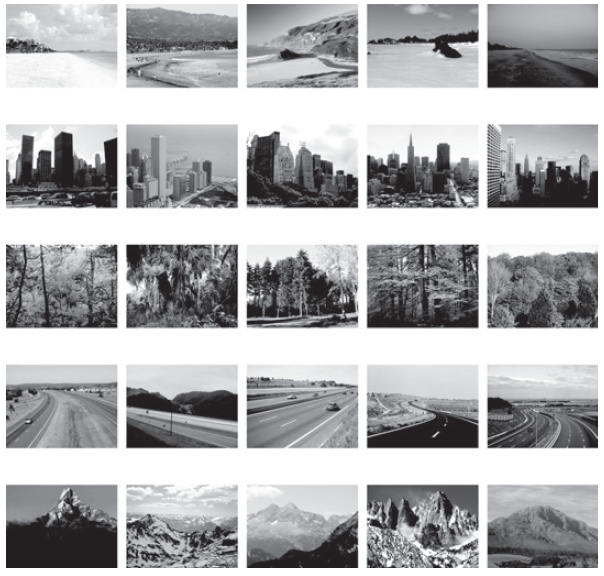}
	
	\includegraphics[width=0.9\linewidth]{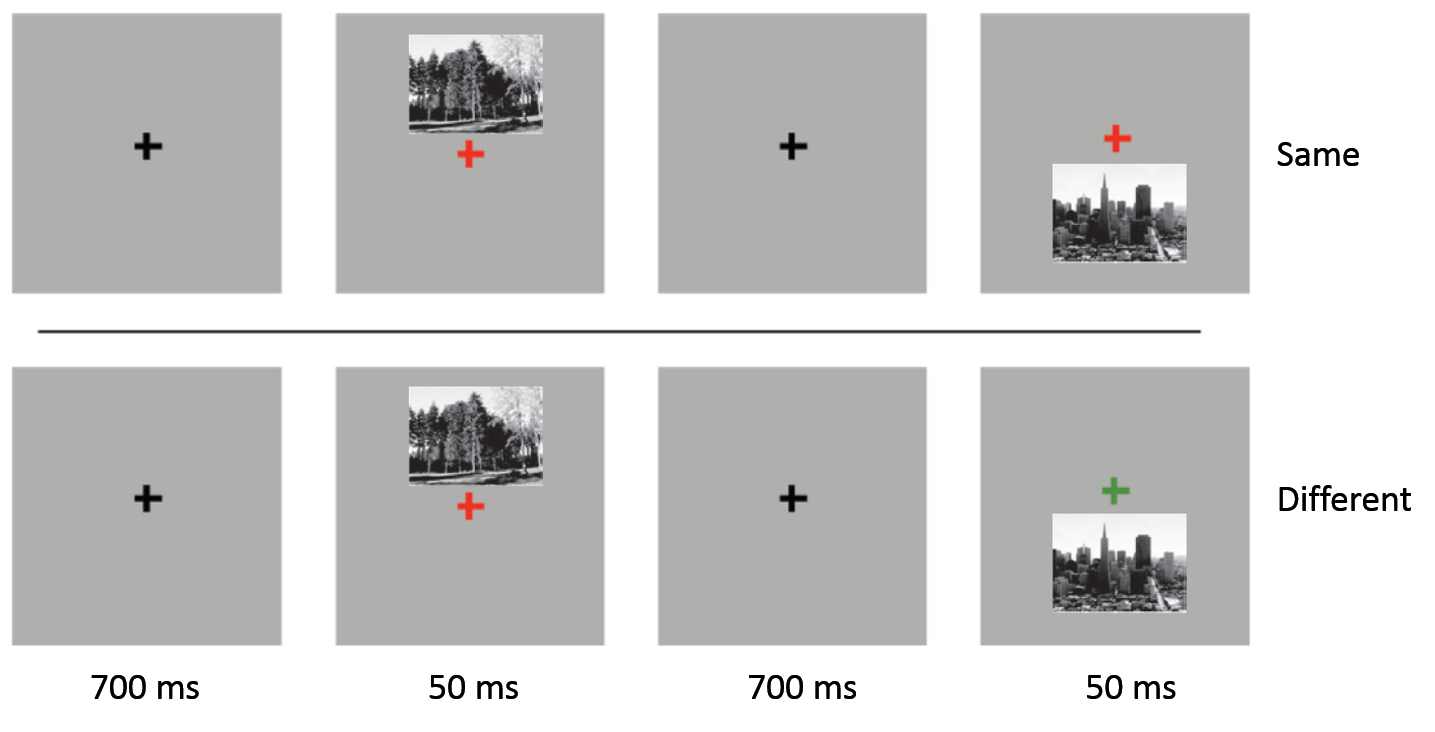}
	\caption{Top: 25 scene images from \cite{walther2009natural} presented in Dataset1, Bottom: experimental task to maintain subject's stare point at the center. The subject is asked to announce if the color of the fixation cross was the same or different at the beginning and end of each two image. The color of the fixation cross are red, blue or green when the images are displayed and  black during inter-stimulus period.}
	\label{Dataset1}
\end{figure}

\subsection{Experimental Data}\label{EXData1}
In this section, we explore the effectiveness of the proposed causal correlation measure on real MEG recordings, with a goal of constructing an activity map of the brain.

\noindent \textit{Dataset 1:}

\noindent This dataset includes MEG signals acquired in the Center for Biological and Computational Learning (CBCL) at MIT \cite{isik2014dynamics}. An Elekta Neuromag Triux with 102 magnetometers and 204 planar gradiometers is used to record the MEG signals.  This dataset is collected on one subject with normal vision  viewing 25 scene images shown in Figure \ref{Dataset1} displayed in the center of the visual field. The images are taken from \cite{walther2009natural} and publicly available at http://vision.stanford.edu/ projects/sceneclassification/resources.html. 
All images were shown in gray-scale and each image was presented for 48 ms with 704 ms inter-stimulus interval. The orderings of shown images were randomly selected, and each stimulus was repeated 50 times resulting in 1250 trials. 
In the data acquisition process, a subject performed a task irrelevant to the images displayed in order to assure the central fixation is continued.
The images were shown with a fixation cross, and the cross changed color randomly between red, blue or green when each image was displayed. The cross changed to black during the inter-stimulus interval, and to another random color when the next image was shown. The subject was required to announce if the color of fixation cross is different at the beginning and end of each image or not (Figure \ref{Dataset1}).

\begin{figure}[tb]
      \centering
      \includegraphics[width=1\linewidth]{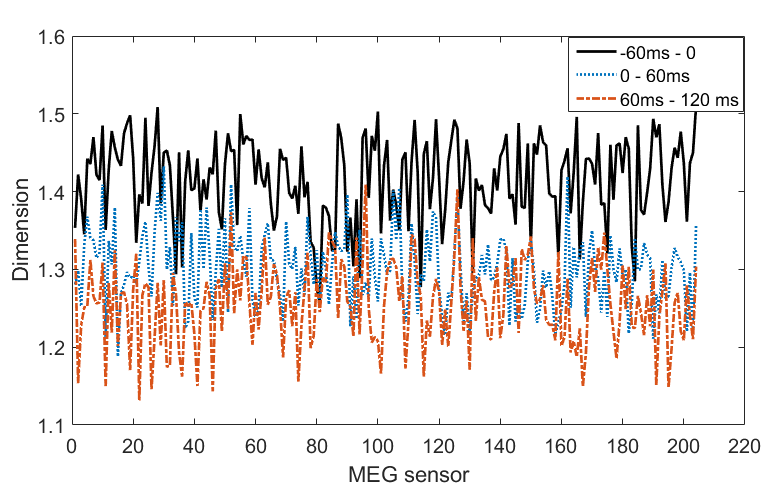}
      \includegraphics[width=0.8\linewidth]{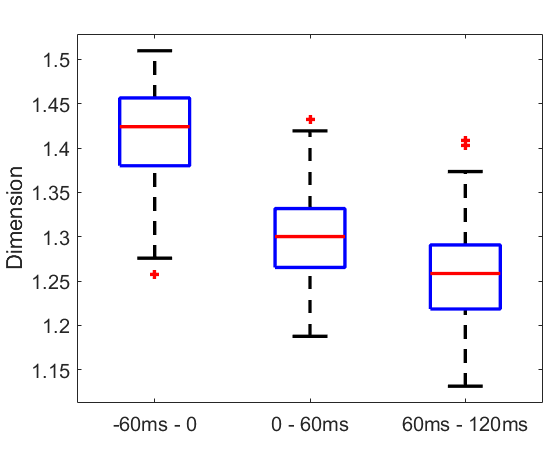}
      \caption{Top: min dimensions of pairwise multivariate delay embeddings for each MEG sensor measurement with respect to the rest of the sensors for three windows of length $60$ms: ($-60$ms - $0$ms), ($0$ - $60$ms) and ($60$ms - $120$ms). Bottom: box plots indicate the distribution of dimensions over 204 for sensors in the three time windows.}
       \label{Dataset1_Dims1}
   \end{figure}  

The MEG data is sampled at 1000 Hz and pre-processed using Brainstorm software \cite{tadel2011brainstorm}. Signal Space Projection method is applied for filtering movement and sensor contamination \cite{tesche1995signal}. Moreover, the MEG signals were band-pass filtered via a linear phase finite impulse response (FIR) filter with lower and higher cut-off frequencies of $2$HZ and $100$Hz in order to eliminate environmental disturbances and unrelated biological noise.

The data from 204 gradiometers are used for this experiment. 
For each sensor measurement $\{x_{k,n}\}_{n=1}^N$, we calculate $d_{kj}$ using equation (\ref{weight}) as described in Section \ref{ConMaps}. 
%
We proceed to find the smallest dimension between all the calculated $d_{kj}$'s for sensor $k $ and the rest of the sensors as
\begin{equation}
d_k = \displaystyle{\min_{j \in \{1,2,...204\}\setminus k} d_{kj}  }.
\end{equation}
This value corresponds to the dimension of the point cloud corresponding to sensor $k$ and the sensor with the strongest causal interaction with sensor $k$. Moreover, $1/d_k$ denotes the strength of this interaction.
Figure \ref{Dataset1_Dims1} shows the values of $d_k$ for MEG sensor indices $k \in \{1,2,...,204 \}$, computed in windows of 60 ms length  including one window before the stimulus onset ($-60$ms - $0$ms) and two windows during the stimulus starting at the onset time ($0$ - $60$ms) and ($60$ms - $120$ms). Clearly, the calculated dimensions for data recorded during the stimulus are lower than those computed during the rest mode (inter-stimulus period). The distributions of dimensions for all the 204 sensors   for ($-60$ms - $0$ms), ($0$ - $60$ms) and ($60$ms - $120$ms) are shown in box plots in Figure \ref{Dataset1_Dims1}. The dimensions in the time window before the onset of experiment are significantly higher than the dimensions in both windows after showing the image $(p < 0.001)$. Moreover, the dimensions in the time window between $60$ms and $120$ms are significantly lower than the dimensions in the first time window after displaying the image ($60$ms - $120$ms) at a p-value $p < 0.001$.
 These results verify that the level of causal interactions in the brain is higher during the visual processing compared to resting. 
Moreover, the dimensions are lower in the second window after stimulus compared to the first one, showing that there are more causal interactions between $60$ms and $120$ms after the onset of stimulus compared to the first  $60$ms of viewing the image. This result is consistent with the findings presented in \cite{isik2014dynamics}, showing strong activity in the visual cortex beginning 60 ms after stimulus onset.
We repeated this experiment on the whole dataset including 51 repetitions of all the 25 images shown in Figure \ref{Dataset1} and have obtained the same results with significance $p < 0.001$. 

\section{MEG-based Decoding of Visual Stimuli using CIM}\label{Decoding}
In this section, we use another dataset of MEG recordings acquired during processing of visual stimuli with the objective of decoding different categories including artificial, natural and soccer game videos on the basis of pairwise causal interactions between 204 brain regions.

\textit{Dataset 2:} This dataset is taken from ICANN/PASCAL2 Challenge ``MEG Mind
Reading'' held in conjunction with the International Conference on Artificial Neural Networks (ICANN) 2011, publicly available at http:// www.cis.hut.fi/ icann2011/meg/measurements.html. This data includes MEG recordings collected with a similar equipment apparatus as Dataset 1 as a result of one subject viewing videos without audio on two different days. The data is provided for 204 planar gradiometer channels.
The raw MEG signals are low-pass filtered and downsampled to a sampling frequency of $200$Hz. The environmental disturbances are removed and head movements are compensated for using the signal space separation (SSS) method proposed in \cite{taulu2006spatiotemporal}. The dataset includes $1$ second long ($200$ samples) recordings measured during stimulus.

The goal of our study is to decode different visual stimuli by recognizing the category of the displayed stimulus. Three different types of stimuli are used in this experiment. The first category is artificial stimuli that include screen savers showing animated artificial objects, shapes or text. The second type is natural stimuli that are video clips from natural documentaries, showing natural sceneries such as mountains and oceans. The third category includes clips taken from soccer matches of Spanish LaLiga.

Since the generalization of brain decoding to new stimuli is crucial, the actual stimulus content needs to be different in training and testing sets while they still lie in the same category. In other words, the goal is not recognizing whether the subject is looking at the exact same image of a specific object but to identify which category of images is observed. Therefore, while part of the training and test sets correspond to the same stimuli, the other part of the test set contains recordings of different stimuli not used in the training set. Specifically, 33\% of the measurements in the test set correspond to the stimuli not displayed in the training set. Moreover, the $1$ second long training and test data are taken from the original long recording with 1 second timing offset. Therefore, 77\% of the measured data with overlapping stimuli between the training and test sets are not recorded at the precise same time, and include different contents.
In addition to generalization to various stimuli of the same category, the decoding task needs to be effective during recording sessions at different times. Accordingly, we use the training and test sets collected during two separate days. The training set includes data acquired in the first day of experiment, and the data in test set is collected in the second day. A small portion of the test set (7\%) is labeled to simulate a very short training period during the test session. This set includes 10 labeled samples in each class recorded in day 2. The number of recordings of artificial, natural, and soccer stimuli in training and test sets are shown in Table 1.
%

%
\begin{figure}[tb]
	\centering
	\includegraphics[width=1\linewidth]{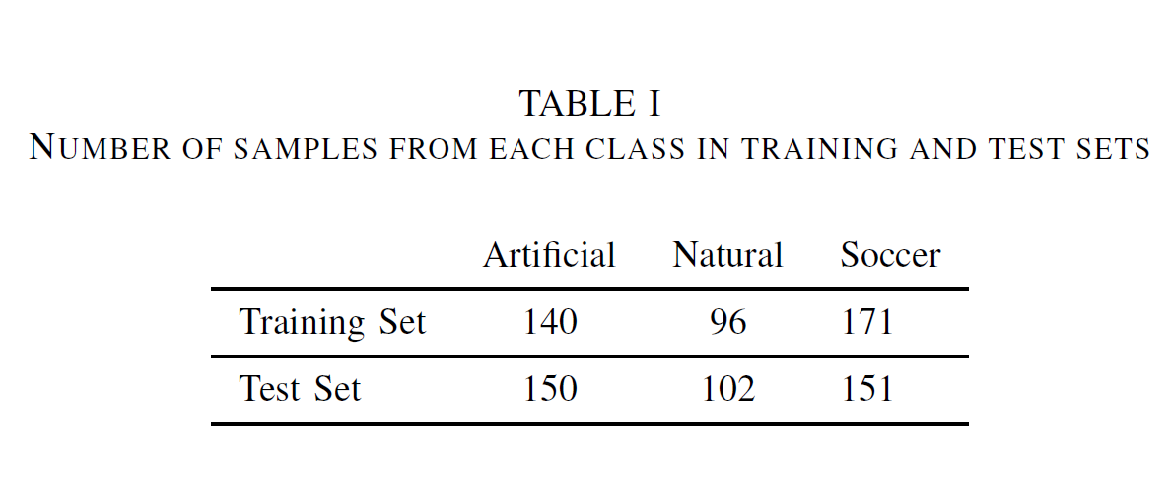}
	
\end{figure}

For each $1$ second of MEG data, we construct the corresponding effective connectivity map using pairwise causal interactions as described in Section \ref{ConMaps}, resulting in a non-symmetric $204 \times 204$ adjacency matrix $\bm{A}$ (Equation (\ref{Adj})). A possibly maximal causality delay of 500 ms is used for this dataset. We use the elements of the adjacency matrix $\bm{A}$  representing the strength of pairwise causal interactions as features for classification. Since the non-symmetric causality matrix would result in 41412 features and the number of available 
observations is much smaller, we use only outgoing edges of the connectivity maps resulting in 20706 features. We then apply a statistical test to reduce the number of significant features. For each feature, we apply a Kruskal–Wallis test to the observations in the training set to determine whether that feature is significantly different between the three classes. The motivation is to identify discriminant features that are effective in differentiating between artificial, natural and soccer stimuli. Kruskal–Wallis test is the non parametric equivalent of one way ANOVA(analysis of variance) on ranks. Therefore, it is useful in this case where we cannot necessarily assume normal distributions. Using $1\%$ significance level for all the causal connectivity features, 358 features are selected to be significantly differentiative. We use these features as inputs to a generalized linear model.
Considering a dataset with $N$ observations and $p$ predictors, the common linear regression model, given the response $\mathbf{Y} =(y_1, \dotsc , y_n)^T$ and the model matrix $\mathbf{X} = ( \mathbf{x}_1 | \dotsc |\mathbf{x}_p)$, with
predictors $\mathbf{x}_j= (x_{1j},\dotsc, x_{nj})^T$, is represented by 
\begin{equation}
\hat{\mathbf{Y}} =\hat{\beta}_0+\mathbf{x}_1 \hat{\beta}_1+ \dotsc \mathbf{x}_p \hat{\beta}_p.
\end{equation}
The objective of a model fitting method is to obtain the coefficients $\hat{\beta}= (\hat{\beta}_0, \hat{\beta}_1, \dotsc, \hat{\beta}_p)$.
Lasso \cite{tibshirani1996regression} is a well-known method for regression that uses an $l_1$ penalty to
obtain a sparse solution. However, if the number of observations is less than the number of predictors $N < p$, the Lasso chooses a maximum of $N$ variables, which is a constraining characteristic for our problem since we do not have access to a large number of observations. Moreover, if there is a set of variables with large pairwise correlations, the Lasso chooses only one variable from the group and ignores the rest of the group resulting in failure in group selection. The elastic net regularization was subsequently proposed in \cite{zou2005regularization} for 
correlated variables using a penalty that is partly $l_1$ and partly $l_2$.
The elastic net solves the following problem
\begin{equation}
\min_{(\beta_0,\beta) \in \mathbb{R}^{p+1}} \left[ \frac{1}{2N} \sum _{i=1}^N (y_i - \beta_0 -\mathbf{x}_i^T \beta)^2 +\lambda P_\alpha(\beta) \right],
\end{equation}
where 
\begin{equation}
P_\alpha (\beta) = \frac{(1-\alpha)}{2} \|\beta\|_2^2 +\alpha \|\beta\|_1
\end{equation}
represents the elastic-net penalty. Elastic-net is the same as Lasso for $\alpha =1$ and becomes ridge regression for $\alpha =0$. Ridge regression narrows the coefficients of correlated predictors close to each other and keeps all the predictors in the model. For $0< \alpha <1$, $P_\alpha$ is a compromise between $l_1$ norm and squared $l_2$ norm of $\beta$.
The $l_1$ part of the penalty term creates a sparse model allowing to choose a reduced set of predictors and the $l_2$ norm removes the constraint on the number of chosen variables, promotes the grouping effect and stabilizes the $l_1$ regularization. Elastic net penalty is advantageous in this study since there are many correlated features in effective connectivity maps, and the pairwise causal interactions cannot be assumed independent from each other, especially in neighboring brain regions. Moreover, we need the $l_1$ term as well since we include the strengths of all pairwise causal interactions as features while not all of them are differentiative between different classes of visual stimuli. For fitting the generalized linear model with elastic-net penalties, we use the algorithm proposed in \cite{friedman2010regularization} for multinomial regression problem. A 10-fold cross validation
is used in the training set to select the model parameter $\lambda$. Using elastic net penalty with $\alpha =0.6$, we could achieve the overall classification accuracy of $\mathbf{77.42 \%}$. This classification accuracy is promising since the best results for classification of these three classes of short clips obtained by the winner of the ICANN/PASCAL2 challenge is $\mathbf{67.5 \%}$. Moreover, we obtained this high accuracy by using only the strengths of pairwise causal interactions between MEG sensor measurements calculated by our proposed measure CIM with no more additional features. 
The confusion matrix of the three classes are shown in Table 2, where the rows correspond to actual classes and the columns are the predicted classes. The accuracy within each category for artificial, natural and soccer stimuli are $79.33\%$, $67.65\%$ and $82.12\%$, respectively. Clearly, the obtained within class accuracy is lowest for the second class that has the smallest number of observations and highest for the class with the largest number of observations (Table 2). Accordingly, we can achieve even better decoding accuracies if we have access to a bigger dataset including more MEG recordings. Considering the fact that the whole training set utilized in this study includes only less than 7 mins of MEG recordings, this objective can be achievable.
\begin{figure}[tb]
	\centering
	\includegraphics[width=0.7\linewidth]{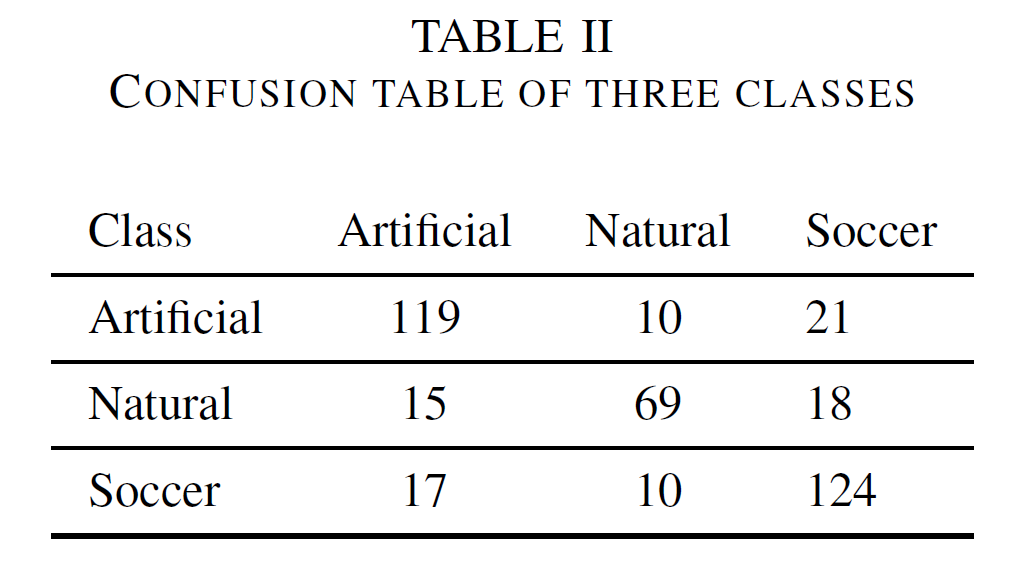}
	
\end{figure}

Note that our reason for selecting Dataset 2 for the decoding task instead of Dataset 1, is due to the presence of natural language processing activities in Dataset 1. This corresponds to the unrelated task of naming the color of the fixation cross. This speaking related activity can be observed in the connectivity maps. However, since we do not know what word (red, blue or green) was pronounced during what trial, we were not able to filter out the language processing activities. On the other hand, Dataset 2 only includes visual activities and hence is appropriate for our objective of decoding visual stimuli.

 The remainder of the paper investigates the evolution of effective connectivity maps by analyzing the toplogical features over varying weights, and shows different geometric patterns in the networks corresponding to visual stimuli with different structures.

\section{Persistent Homology of Effective Connectivity Maps for Detecting Geometric Structures }\label{PHEC}
\subsection{Proposed Method}\label{PHECMethod}
Effective connectivity maps are highly complex, completely connected, weighted and directed networks. Depending on whether we threshold the causal interaction values at a determined level or not, we obtain either weighted or binary graphs. Although most of graph analysis measures can be generalized to the weighted networks \cite{opsahl2010node, opsahl2008prominence, newman2001scientific, rubinov2010complex}, simultaneous presence of fully connectedness and weighted edges in the network yields considerable difficulty to interpret brain connectivity maps and calls for the broad application of ad-hoc thresholding techniques \cite{sepulcre2010organization, power2011functional, ginestet2011brain, he2008structural, van2010comparing}. 
%
%
The structure of the network heavily dependent on the selected threshold makes it challenging in disregarding edges with lower weights, which results in risk of trade-off between information comprehensiveness and simplicity. It also jeopardizes neglecting the potential influence of connections with small weights in the network. This effect has been studied in resting state functional connectivity \cite{schwarz2011negative}, prediction of cognitivve control \cite{cole2012global}, and showing the importance of weak pairwise correlations in strong correlated network states in neural populations\cite{schneidman2006weak}.
To select the appropriate threshold in the brain networks, multiple comparison correction over all possible connections is proposed \cite{van2010comparing, rubinov2009small, ferrarini2009hierarchical, bohland2009brain}. However, the structure of the final network heavily relies on the the selected $p$-value of thresholding.
One way of approaching this problem is by controlling the network density defined as the proportion of the number of existing connections to the number of maximum possible edges \cite{van2010comparing, achard2007efficiency, van2009efficiency}. This approach needs an expert estimation and a pre-determined range of density and is also dependent on the chosen sparsity level \cite{bassett2006adaptive, achard2007efficiency, he2008structural}. 
There is no broadly established principle for ad-hoc thresholding to date. 
To address this issue, an alternative to threshold selection is considering a set of networks for every possible threshold and analyzing the evolution of network changes over various thresholds \cite{lee2012persistent}. In other words, we simultaneously analyze the network connectivity structures and weights for each threshold, and analyze how they vary with the threshold. Since we analyze connectivity maps built at all possible thresholds, we avoid the issue of finding a suitable threshold.

Persistent homology provides a tool for efficient analysis of considerably many networks by evaluation of their topological features over varying scales \cite{Hatcher2002, Edel10, Zomorodian2004}. We first briefly explain the necessary concepts from algebraic topology and then describe how persistent homology will be applied to effective connectivity maps. More details about the theory of persistent homology and its computation is available in \cite{edels2002, carlsson2009topology, cohen2007stability}.  
Consider a set of vertices $V$. A simplex of dimension $d$ is defined as 
the convex hull of $d+1$ nodes $[v_1,v_2, ..., v_{d+1}]$. For example, a simplex of dimension 0, 1, 2, 3 is respectively a vertex, edge, filled in triangle and pyramid. A simplicial complex is a union of simplices $K=\{\sigma_1,\sigma_2,...\sigma_n\}$ such that for any two simplices $\sigma_i,\sigma_j$ we have either $\sigma_i\cap \sigma_j=\emptyset$ or $\sigma_i\cap\sigma_j=\sigma_k$, where $\sigma_k\inn K$.
Given a simplicial complex $K$, we define an ordered family of vector spaces as follows. Let $C_i(K)$ be the abstract vector space created over a field $\mathbb F$ such that each simplex of dimension $i$ is a basis element. Clearly, if the dimension of the complex is $d$, we have $d+1$ vector spaces created, namely $C_0(K),C_1(K), ... C_d(K)$. Then one can define the so-called boundary maps $\partial_i:C_n(K)\to C_{n-1}(K)$ for each simplex as:
\begin{equation}
 \partial([v_0,v_1, ..., v_n])=\Sigma_{i=0}^n (-1)^i[v_0,v_1, ...,\widehat{v_i} ..., v_n].
\end{equation}
It is easy to check that the composition of two subsequent boundary maps is always zero, i.e.  $\partial_i\circ\partial_{i-1}=0$ for all $i>0$. Those boundary operators can be extended to the whole $C_i(K)$ by linearity. Thus one can obtain a chain of vector spaces and linear maps to form a chain complex as \cite{Hatcher2002}:
\begin{multline}
0\to C_d(K)\stackrel{\partial_d}{\to}C_{d-1}(K)\stackrel{\partial_{d-1}}{\to} ...\\ ... \stackrel{\partial_{3}}{\to}C_2(K)\stackrel{\partial_{2}}{\to} C_1(K)\stackrel{\partial_{1}}{\to}C_0\to 0.
\end{multline}

\noindent The space $H_n(K)=\displaystyle{\frac{Ker \partial_n}{Im\partial_{n+1}}}$ is called the $n$-th dimensional homology of the space $K$. The dimension of this space is denoted by
\begin{equation}\label{BettiDef}
 \beta_n=dim(H_n(K)),
\end{equation}
and is called the $n$-th Betti number of the space $K$, and completely characterizes the $n$-th dimensional homology of the space $K$.
The $n$-th Betti number intuitively counts the number of different $n$-dimensional topological features of the space $K$. For example, $\beta_0$ measures the number of connected components, $\beta_1$ counts the number of ``loops", $\beta_0$ represents the number of ``voids".



In an effective connectivity map with a set of nodes $V$, for each threshold on the weights $\epsilon$, we obtain a network $ \mathfrak{B} (V,\epsilon)$ represented by a binary adjacency matrix $\bm{A}^\epsilon = [a^\epsilon_{ij}]$, where
\begin{equation}
 a^\epsilon_{ij} = \left\lbrace  \begin{array}{ccc}
                               a_{ij} &  & a_{ij} \leq \epsilon ,   \\
                               0 &  & \text{otherwise}
                             \end{array}
                             \right.
\end{equation}
Starting with an initial threshold $\epsilon_0$ and increasing it as $ \epsilon_0 \leq \epsilon_1 \leq \cdots \leq \epsilon_n$, we obtain a nested sequence of graphs as
\begin{equation}
\mathfrak{B} (V,\epsilon_0),\mathfrak{B} (V,\epsilon_1),\mathfrak{B} (V,\epsilon_2),\cdots, \mathfrak{B} (V,\epsilon_n).
\end{equation}
In order to have a comprehensive representation, we start with the smallest weight in the effective connectivity map, and each subsequent
graph includes an additional edge $ij$ corresponding to the
next greater entry of the adjacency matrix $\bm{A}$. This procedure will result in a sequence of nested binary networks, namely a graph filtration as:
\begin{equation}
\mathfrak{B} (V,\epsilon_0)\subseteq \mathfrak{B} (V,\epsilon_1)\subseteq\cdots \subseteq \mathfrak{B} (V,\epsilon_i) \subseteq \cdots \subseteq \mathfrak{B} (V,\epsilon_n),
\end{equation}
for $ \epsilon_0 \leq \epsilon_1 \leq \cdots \leq \epsilon_n$.
To each of these graphs $\mathfrak{B} (V,\epsilon)$, we associate a flag complex $K_\epsilon$, which is the simplicial complex that includes the $k$-simplex $[v_1, v_2, \cdots v_{k-1}]$ whenever the nodes $v_1, v_2, \cdots v_k$ form a clique. A clique is defined as a completely connected subgraph contained in an unweighted undirected graph. 
A sequence
of topological spaces $X = X_0, X_1, \cdots, X_N$ form a filtration if $X_i \subseteq X_i$ whenever $i<j$. Therefore, the series of complexes $\{K_\epsilon \}$ form a filtration since  $K_{\epsilon_i} \subseteq K_{\epsilon_j} $ for $\epsilon_i<\epsilon_j.$
For increasing values of threshold $\epsilon$, one gets a nested sequence of simplicial complexes as
\begin{equation}\label{nestedcplx}
 K_{\epsilon_0}\subseteq K_{\epsilon_1} \subseteq ... \subseteq K_{\epsilon_i} \subseteq ... \subseteq K_{\epsilon_n}. 
\end{equation}
\noindent For each threshold, the simplices in $K_{\epsilon_i}$ derive their structure from the corresponding network  $\mathfrak{B} (V,\epsilon_i)$. 

 \begin{figure}[tb]
      \centering
      \includegraphics[width=1\linewidth]{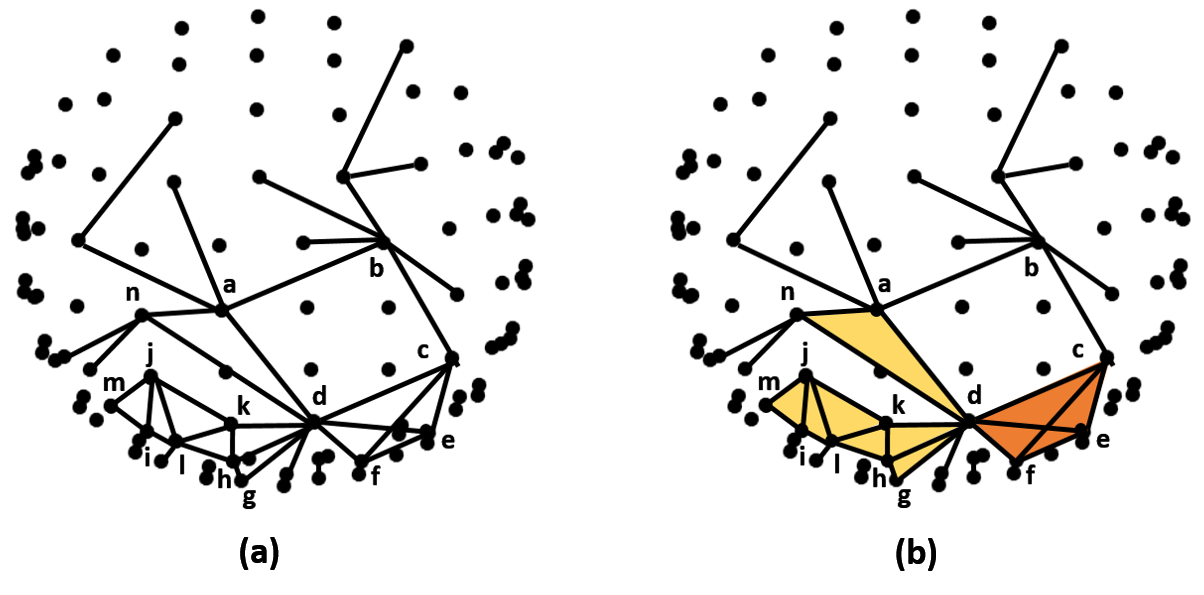}
      \caption{(a) An unweighted network with areas of dense and sparse connectivities. The cycle $(a, b, c, d)$ forms a one dimensional hole, representing a sparse connectivity region, while the 3-cliques and 4-clique $(c,d,f,e)$ represent dense regions in the network.
    (b) The associated flag complex achieved by upgrading cliques to simplices. Cliques of size 3 and 4 are shown with yellow and orange colors, respectively.}
       \label{Cliques}
   \end{figure} 
 \begin{figure}[tb]
 	\centering
 	\includegraphics[width=1\linewidth]{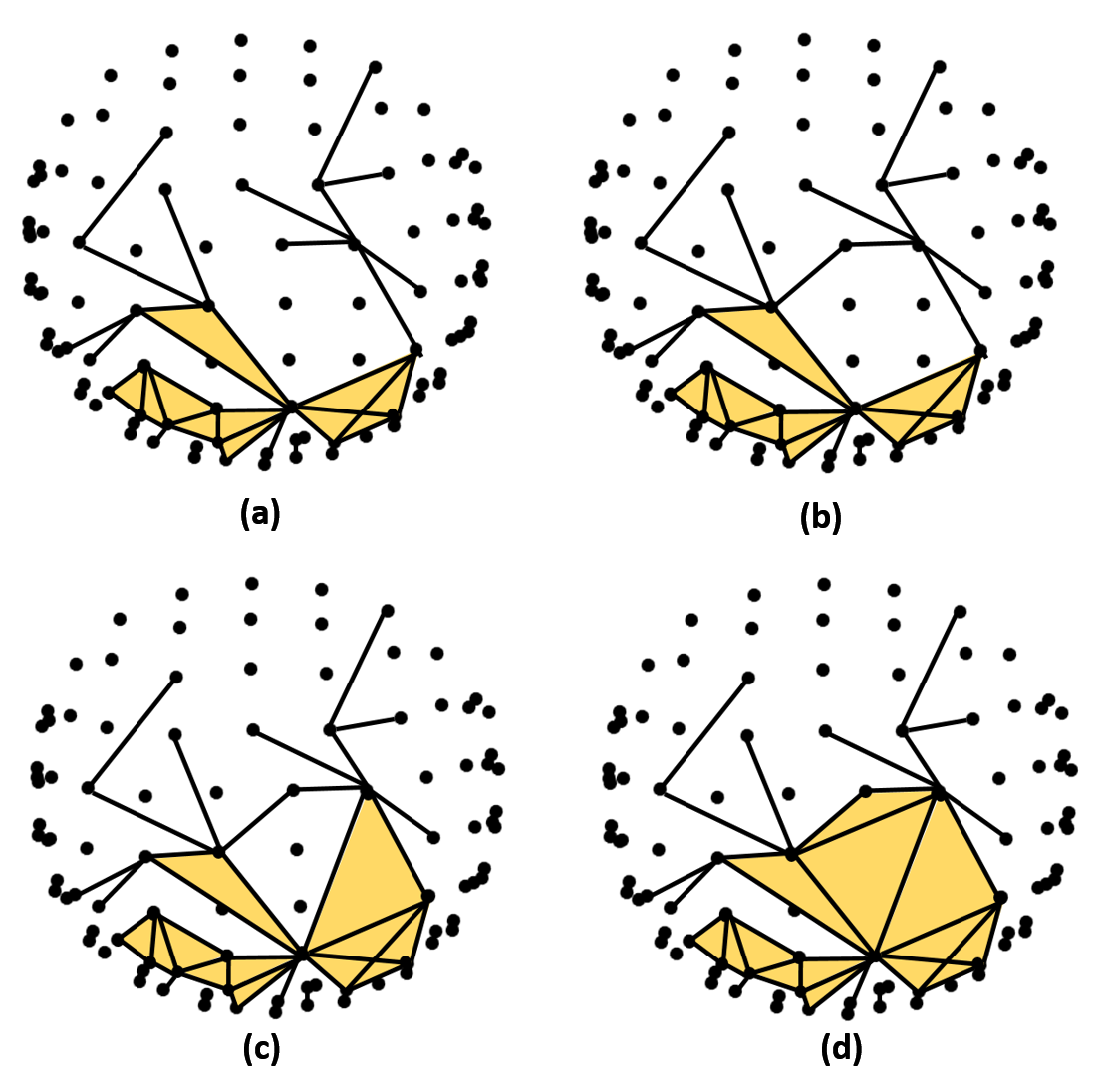}
 	\caption{The weight filtration: a cycle appears, shrinks and dies as the threshold level grows. (a) At initial threshold value $\epsilon_0$, there is no cycle. (b) As the threshold increases to $\epsilon_1$, a new one dimensional cycle is formed. (c) The cycle persists while getting smaller at threshold $\epsilon_2$ and eventually dies as it gets filled by cliques (d). 
 	}
 	\label{Filtration}
 \end{figure}

Figure \ref{Cliques} (a) shows an unweighted network where the cycle $(a, b, c, d)$ forms a one dimensional hole. This hole represents a sparser region with lower connectivity compared to the neighboring areas. On the other hand, the cycles of three nodes and the fully connected 4-cycle $(c,d,f,e)$ represent dense regions in the network. Since cliques are used for determining the density of networks, it is appropriate to represent the three node cycles and 4-cycle $(c,d,f,e)$ with 3-cliques and 4-cliques, respectively. In Figure \ref{Cliques} (b), the cliques are filled and the decreased connectivity region is still present in $(a,b,c,d)$. Since a $k$-clique can be viewed as a $(k-1)$-simplex, the connectivity network can be represented by a flag complex shown in Figure \ref{Cliques} (b).

One important characteristic of cliques is that of being nested, i.e. for a fixed set of nodes, a clique of dimenion $j$ ($j$-cliques) comprises all the cliques of dimension $i$ ($i$-cliques) for $\forall i <j$. This helps construct a filtration by filling the cliques. As the threshold $\epsilon$ increases, the filtration sweeps over all weight scales, the connectivity maps expand, and as a result the topological characteristics of the corresponding flag complexes change. The evolution of connectivity maps over all thresholds provides comprehensive information about the network structure that can be captured by topological features of the associated complexes. 
These topological features can be quantified using Betti numbers. As the threshold increases, new holes are born, modified, and finally die. Figure \ref{Filtration} shows this weight filtration, where a cycle appears, shrinks and dies as the threshold level grows.
These changes can be pursued by measuring topological features of the arrangement of cliques that is referred to as clique topology \cite {giusti2015clique}. The information in the clique topology is quantified
by counting the number of holes or non-tractable cycles after all cliques have been filled. The dimension of the hole and the dimension of its boundary are closely related. $n$-dimensional holes are bounded by $(n-1)$-dimensional faces. For example, the boundary of a two dimensional hole is a one dimensional loop, and a three dimensional void is bounded by a two dimensional surface. Since $\beta_n$ counts the number of $n$-dimensional holes, clique topology is represented by a sequence of Betti numbers $\beta_n (K_\epsilon)$ calculated in different dimensions $i=0,1,2, \cdots$ for the flag complex $K_\epsilon$ associated to each graph $\mathfrak{B} (V,\epsilon)$. In other words $\beta_n$ counts the number of different $n$-cycles in each network after all cliques are filled.

\begin{figure}[tb]
 	\centering
 	\includegraphics[width=1\linewidth]{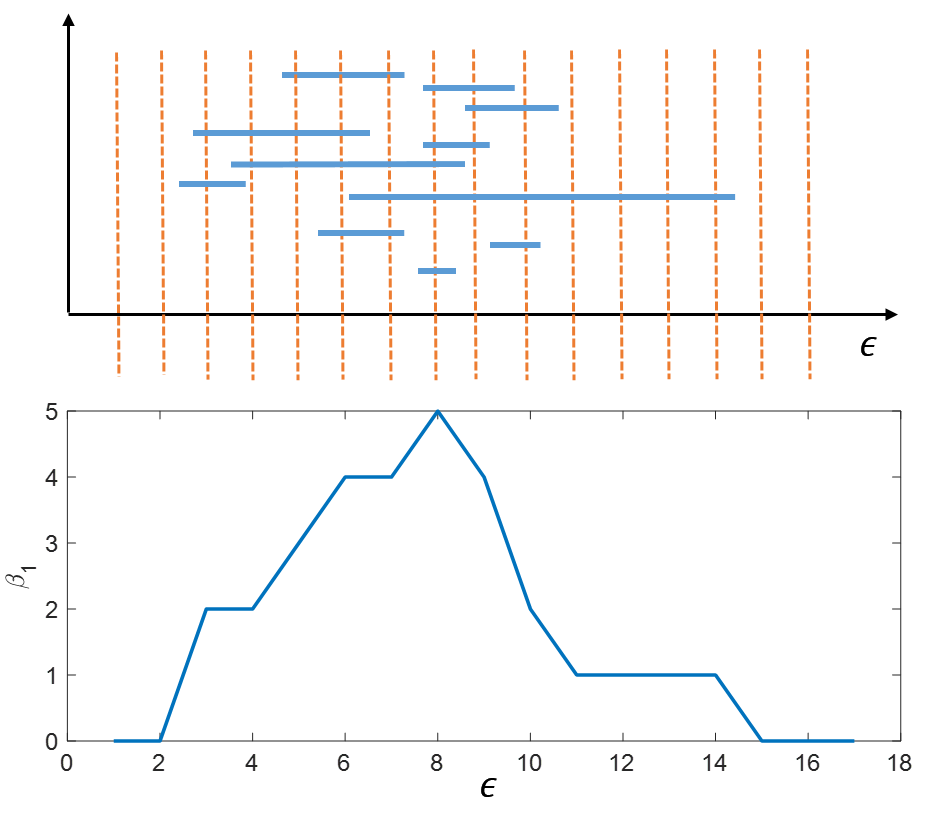}
 	\caption{Top: an example barcode. Bottom: the distribution of its persistence intervals by counting the number of bars present at each filtration value}
 	\label{BettiTraj}
 \end{figure}

The topological changes of the filtration can be visualized using persistence barcodes, which are formed by computing the varying topological features across different filtration values. Consider the nested sequence of simplicial complexes in Equation (\ref{nestedcplx}) and let $H_n^i=H_n(K_i)$.  From the functorial properties of homology one gets a sequence of the form: 
\begin{equation}
H_n^1\to H_n^2\to \dots H_n^i\to \dots\to H_n^k.
\end{equation}
A particular class $\alpha$ may come into existence in $H_n(K_i)$, and it then either gets mapped to zero in $H_n(K_s)$ for some $s>i$ or to a non zero element in the last homology $H_n(K_k)$.  This yields a barcode, a collection of interval graphs lying above an axis parameterized by $\epsilon$. An interval of the form $[t,s]$ corresponds to a class that appears (is born) in $H^t_n$ and is mapped to zero (dies) at $H^s_n$ (Figure \ref{BettiTraj} (top)).

In our previous study on distinguishing harmonic structures from random patterns in signals, we proposed finding the distribution of persistence intervals in each barcode by calculating the number of bars
present at each filtration radius in order to find an appropriate threshold \cite{meicassp14}. In this study, we use a similar framework for the barcodes corresponding to each Betti number $\beta_n$ by computing the number of bars
present at each threshold. The number of bars calculated over all graphs in the filtration in dimension $n$ is represented by $\beta_n(\epsilon) = \{\beta_n(\epsilon_r)\}_{r=0}^n$ , where $\epsilon_0$ is the smallest connection weight in the network $\epsilon_0 = \min(a_{ij}), \forall i,j$. We refer to $\beta_n(\epsilon)$ as $\beta_n$ trajectory. Figure \ref{BettiTraj} shows an example barcode and its corresponding $\beta_1$ trajectory obtained by counting the number of bars present at each filtration value.
In analyzing effective connectivity maps, the structure of each single network corresponding to each particular threshold might be affected by noise, clique topology however provides a robust framework for analyzing the evolution of the network. 

The arrangement of cliques and how they evolve by increasing threshold provides information about hidden geometric structures of the network.  In the networks with geometric structures, cliques are more dominant compared to the random networks. Therefore, the holes in more geometric networks are less persistent since cliques fill in the holes quicker compared to random networks.
 In \cite{giusti2015clique}, clique topology has been independently used to distinguish random and geometric structures in neural connectivity networks constructed using pairwise correlation of rat hippocampal place cell activities. The utilized data in \cite{giusti2015clique} was collected inavasively while the animal was exploring a 2D environment. In this study, we analyze the evolution of effective connectivity maps obtained by our proposed causal interaction measure (CIM) using MEG data recorded during visual stimuli. We calculate the number of one dimensional holes(cycles) over all graphs in the filtration as the threshold on weights increases. Comparing different visual stimuli, the Betti trajectories of effective connectivity maps reveal that the networks created for the collected data while the subject was viewing more ``structured" images are more geometrically organized, as will be explained in the next Section.
 

\subsection{Experimental Results }
 \begin{figure}[tb]
      \centering
      \includegraphics[width=1\linewidth]{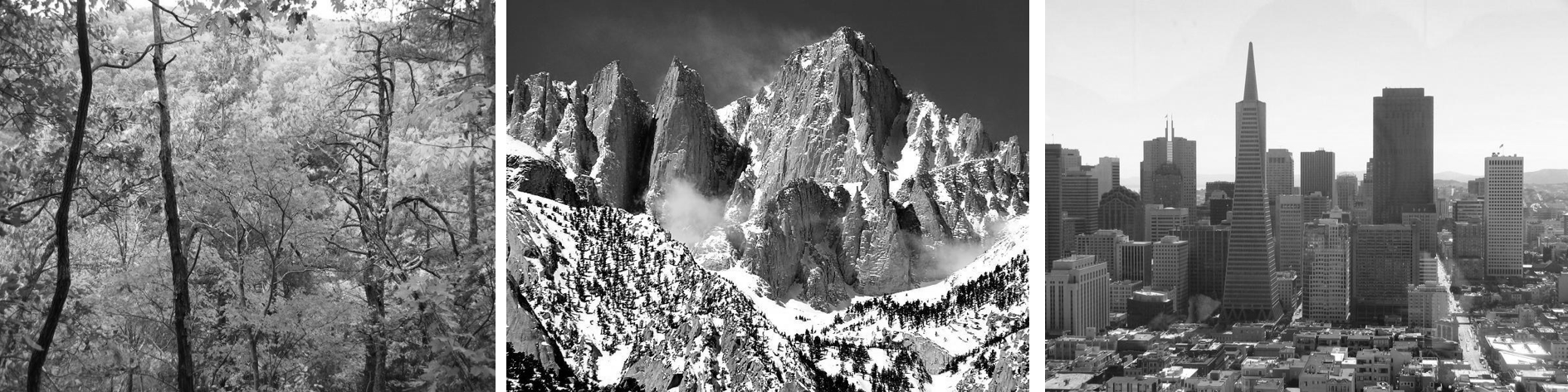}
      \caption{Scene images of Forest, Mountain and Buildings used as visual stimuli of different geometric structures }
       \label{FMB}
   \end{figure}

    \begin{figure*}[tb]
      \centering
      \includegraphics[width=0.9\linewidth]{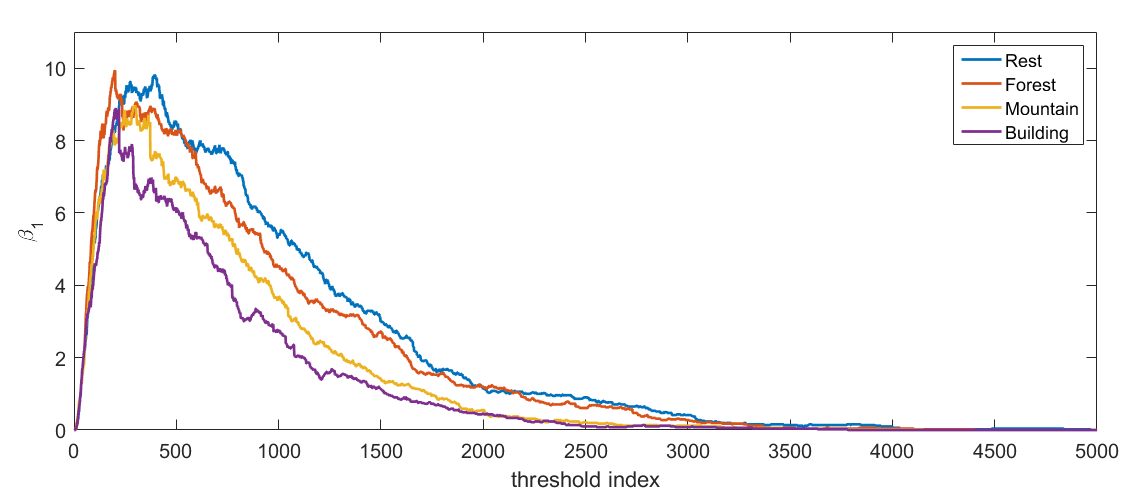}
      \caption{Mean Betti trajectories $\beta_1(i)$ with respect to threshold index $i$ shown for effective connectivity maps constructed using CIM measure for MEG data recorded while the subject was viewing images of Forest, Mountain and Building shown in Figure (\ref{FMB}) in purple, yellow and orange, respectively. The results are also shown for the rest mode when no image was displayed (blue).  }
       \label{Betti1Curves}
   \end{figure*} 

In this section, we apply the topological framework described in Section \ref{PHECMethod} to effective connectivity maps obtained as described in Section \ref{ConMaps} for Dataset 1. We specifically focus on images of Forest, Mountain and Building as shown in Figure (\ref{FMB}). We construct the connectivity maps using our proposed causal interaction measure applied to 204 MEG sensor measurements collected while the subject was looking at each image. Since each image was presented 51 times, we obtain a set of 51 effective connectivity maps for each of the three images.

Since different connectivity maps have different ranges of weights, i.e. the maximum and minimum weights in various networks are not necessarily the same, we sort the connection weights in descending order $ \epsilon_0 \leq \epsilon_1 \leq \cdots \leq \epsilon_n$ and use the ranks $0, 1, \cdots, n$ as threshold indices, where $n=\frac{N(N-1)}{2}$ for undirected graphs with $N$ nodes. Therefore, the first graph in the filtration has one connection with weight $\epsilon_0$, where $\epsilon_0  = \underset{i,j}{\min} (a_{ij}),$ and as we proceed in the filtration in each step one connection with the next highest weight will be added to the network. This technique helps normalize the range of weights used in the filtration, and aligns the threshold levels of different networks for comparison. For each graph $\mathfrak{B} (V,\epsilon_i)$ in the filtration, we calculate the dimension of the homology group $H_1(K_{\epsilon_i})$ as $\beta_1 = dim(H_1(K_{\epsilon_i}))$ (see Equation (\ref{BettiDef})), where $K_{\epsilon_i}$ is the flag complex associated with graph $\mathfrak{B} (V,\epsilon_i)$. We implement this by counting the number of bars for each filtration value or threshold index in the barcode corresponding to $\beta_1$. This will yield Betti trajectory $\beta_1(i)$ as a function of threshold index $ 0 \leq i \leq n$.

We calculate $\beta_1$ for each threshold index resulting in a trajectory for each repetition of each image in Figure (\ref{FMB}).  Figure (\ref{Betti1Curves}) shows the Betti trajectories $\beta_1(i)$ of effective connectivity maps corresponding to the shown images of ``Forest", ``Mountain" and ``Building" averaged over different repetitions of the same image. We also perform this experiment for the data collected during inter-stimulus period where there was no visual stimulus present with the subject in a rest mode between sessions of viewing images. Clearly, mean Betti trajectories of the connectivity maps for the duration when the subject is viewing images drop faster in comparison to the mean Betti trajectories of the connectivity map in the rest mode. Moreover, the trajectory corresponding to the image of ``Building" clearly drops faster than the one corresponding to the ``Mountain" and the ``Mountain" Betti trajectory drops faster than the ``Forest" trajectory. 

To quantify the difference in these result, we use integrated Betti numbers over all threshold indices as 
\begin{equation}\label{IntegB}
\bar{\beta_1} = \int_{0}^{n} \beta_1(i) dx.
\end{equation}
\noindent It is clear in Figure (\ref{Betti1Curves}) that $\bar{\beta_1}_ {\textit{,Building}} < \bar{\beta_1}_ {\textit{,Mountain}} < \bar{\beta_1}_ {\textit{,Forest}} < \bar{\beta_1}_ {\textit{,Rest}}$. We assessed significance of this result using a bootstrapping statistical test. 
To perform this test, we randomly sampled 5 connecticty maps out of 51 maps available for each image 200 times. In each run, the Betti trajectories of the 5 trials are averaged to generate mean Betti trajectories. This full procedure is repeated 200 times using randomly sampled labels verifying that these results are significant with p-value less than $10^{-40}$. 

The obtained results are owed to the fact that 
the arrangement of cliques provide information about the underlying geometry of the network. In networks with more geometric structures, cliques are more dominant. This causes the holes in the network to get filled faster by the cliques and are less persistent in the filtration compared to random networks. By looking at images used in this experiment in Figure (\ref{FMB}), it is obvious that the image of Buildings is more structured compared to the image of Mountains, and the image of Mountains is more structured than that of Forest. This yields effective connectivity map constructed for Building viewing to be more geometric in comparison to the connectivity maps for Mountain viewing. The same goes for the networks of Mountains and Forest, as 
%
 the image of Mountain is more structured than Forest. Moreover, the effective connectivity map generated for MEG data when the subject is in a rest mode is less geometric than that corresponding to active visual stimuli. In other words, the connectivity network has the highest randomness and the lowest geometric pattern when looking at blank board. In summary, viewing more structured images causes the effective connectivity networks to have more geometric patterns. In more geometric networks, cliques are more prominent and fill the holes more rapidly, resulting in less persistent holes compared to random networks.  This can be detected by analyzing Betti 1 trajectories and comparing integrated Betti values in Equation \ref{IntegB}. $\bar{\beta_1}$ of effective connectivity maps constructed using MEG data recorded with more structured visual stimuli, are smaller than those corresponding to less structured and more random stimuli.

\section{Conclusion}\label{Conclu}
In this study, a novel measure for causal interaction using fractal dimension of multivariate delay embedding is proposed and it demonstrates a capacity of detecting linear and nonlinear interactions with no prior model or information about the system. This measure is used to construct effective connectivity maps using MEG data recorded in 204 locations on the brain while viewing visual stimuli. Different categories of visual stimuli including artificial, natural and soccer matches are successfully decoded using the strength of connections in the effective connectivity maps. The obtained decoding results outperform the most accurate prior studies that are based on functional connectivity using correlation, mutual information, etc. Furthermore, by evaluation of topological features of effective connectivity maps over varying scales, we discovered that the networks corresponding to more structured visual stimuli are more geometric, and clique toplogy can be used to distinguish them. The future research direction is towards implementing the introduced frameworks on EEG data as the acquisition of the latter is more practical than MEG. 


%

\appendices
\section{Proof of Theorem \ref{thm1}}

\begin{proof}

	We will substantiate this Theorem using the concept of mutual information. In \cite{kraskov2004}, the authors estimate the mutual information of two variables $X=\{x_i \}_{i=1}^N$ and $Y=\{y_i \}_{i=1}^N$ of dimensions $d_X$ and $d_Y$ using the following well known relation,
	\begin{equation}\label{MI}
	I(X;Y)= H(X)+H(Y)-H(X,Y),
	\end{equation}
	where the entropies $H(.)$ are estimated using $K$-nearest neighbor distances. The joint entropy was first estimated considering the joint variable $(X,Y)$ as 
	\begin{multline}\label{HXY}
	\hat{H}(X,Y)= -\psi(k)+\psi(N)+\log(c_{d_X}c_{d_Y}) \\ + \frac{d_X+d_Y}{N} \sum_{i=1}^N \epsilon_{xy}(i),
	\end{multline} 
	where $\epsilon_{xy}(i)$ is twice the distance of the point $(x_i,y_{i-\tau_1})$ to its $K$-nearest neighbor in the delay embedding point cloud $\{x_n,y_{n-\tau_1}\}$.  $\psi(.)$ is the di-gamma function defined as $\psi(x)=\frac{d}{dx} \log \Gamma (x)= \frac{\Gamma ^ \prime (x)}{\Gamma(x)}$, where $\Gamma$ is the gamma function. $c_{d_X}$ is the volume of the $d_X$-dimensional unit ball that equals $1$ for the maximum norm and $\pi^{d_X/2}/\Gamma(d_X/2+1)$ for the Euclidean norm. $c_{d_Y}$ is defined in the same way by replacing $d_X$ by $d_Y$. Moreover, the entropy $H(X)$ is estimated using the following equation. 
	\begin{multline}\label{HX}
	\hat{H}(X)= -\frac{1}{N} \sum_{i=1}^N \psi [n_x(i)+1] +\psi(N)+\log c_{d_X} \\ + \frac{d_X}{N} \sum_{i=1}^N \log \epsilon_{xy}(i),
	\end{multline}
	where $n_x(i)$ is the number of points whose distance from $x_i$ is less than $\epsilon_{xy}(i)/2$. The estimation for $H(Y)$ is obtained by replacing $X$ by $Y$ everywhere in Equation (\ref{HX}) \cite{kraskov2004}. The mutual information is then obtained from Equations (\ref{MI})-(\ref{HX}) as follows:
	\begin{multline}\label{IXY}
	\hat{I}(X;Y) =  \psi(k) +\psi (N)- \frac{1}{N} \sum_{i=1}^N \psi [n_x(i)+1] \\ 
	-\frac{1}{N} \sum_{i=1}^N \psi [n_y(i)+1] 
	\end{multline}
	This estimation has a bias that builds mainly upon the dimension of the vector variables. However, due to the nature of multivariate delay embeddings setting, we can use conditional mutual information that is shown in \cite{vlachos2010nonuniform} that results in a smaller bias. 
	%
	%
	Conditional mutual information is defined as
	\begin{multline}\label{CMI}
	I(X;Y|Z) = I(X;(Y,Z)) - I(X;Z) \\
	=H(X,Z)+H(Y,Z)-H(Z)-H(X,Y,Z).
	\end{multline}
	The joint entropy $H(X,Y,Z)$ can be estimated using the $K$-nearest neighbors similar to Equation (\ref{HXY}). The entropies $H(X,Y)$, $H(Y,Z)$ and $H(Z)$ can be estimated by projection of the joint space $(X,Y,Z)$ on the suspaces $(X,Y)$, $(Y,Z)$ and $Z$, respectively.
	%
	In this setting, instead of estimating $I(X,Z)$ independently from Equation (\ref{IXY}), we use the projection of the space $(X,Y,Z)$  onto the subspace $(X,Z)$ and apply Equations (\ref{MI}) and the estimation of the projected entropy in Equation (\ref{HX}) to obtain the following projected mutual information \cite{vlachos2010nonuniform}.
	\begin{multline}\label{EsMI1}
	\hat{I}_p(X;Z)= \psi(N)- \frac{1}{N} \sum_{i=1}^N \psi [n_{x}(i)+1] \\ 
	-\frac{1}{N} \sum_{i=1}^N \psi [n_{z}(i)+1]+ \frac{1}{N} \sum_{i=1}^N \psi [n_{xz}(i)+1],
	\end{multline}
	where $n_{xz}(i)$ is the number of points in $(X,Z)$ space whose distance from the $(x_i, z_{i-\tau_2})$ is less than $\epsilon_{xyz}(i)/2$ and $\epsilon_{xyz}(i)$ is twice the distance of the $i$th point in the joint $(X,Y,Z)$ space, i.e. $(x_i, y_{i-\tau_1}, z_{i-\tau_2})$ to its $K$-nearest neighbor. Similarly, the projected mutual entropy for the multivariate delay embedding of $X$ and $Y$ can be represented as:
	\begin{multline}\label{EsMI2}
	\hat{I}_p(X;Y)= \psi(N)- \frac{1}{N} \sum_{i=1}^N \psi [n_{x}(i)+1] \\ 
	-\frac{1}{N} \sum_{i=1}^N \psi [n_{y}(i)+1]+ \frac{1}{N} \sum_{i=1}^N \psi [n_{xy}(i)+1].
	\end{multline}
	We use Equations (\ref{EsMI1}-\ref{EsMI2}) to prove Theorem \ref{thm1}.
	
	According to the box-counting definition of dimension in Equation (\ref{Box}), if the dimension of the point cloud $(x_n,y_{n-\tau_1})$ is lower than the dimension of the point cloud $(x_n,z_{n-\tau_2})$, the number of hyperboxes of a constant size $r$ (squares in this case) required to cover the points in $(X,Y)$ space is going to be lower compared to the ones required to cover the points in $(X,Z)$ space.  Note that the total number of points in the two spaces are the same since they are projections of points in $(X,Y,Z)$ space into the $(X,Z)$ and $(X,Y)$ spaces. Accordingly, the average number of points around each point $(x_i, y_{i-\tau_1})$ in the joint $(X,Y)$ space is higher compared to the average number of points around each point $(x_i, z_{i-\tau_2})$ in the $(X,Z)$ space. As a result, the following equation holds
		\begin{equation}\label{AvgN}
		\sum_{i=1}^N  [n_{xy}(i)+1] > \sum_{i=1}^N  [n_{xz}(i)+1].
		\end{equation}
	To compute the left side of the equation above, for each point $(x_i, y_{i-\tau_1})$, we calculate the number of points contained in a ball of radius $\epsilon_{xyz}(i)/2$ centered at that point plus one. We then average this value for all the points in the joint $(X,Y)$ space. The right part of the equation is obtained using the same procedure on the points $(x_i, z_{i-\tau_2})$ in the $(X,Z)$ space. 
	Since digamma function is strictly increasing, we can conclude that 
		\begin{equation}
		\psi(\sum_{i=1}^N  [n_{xy}(i)+1]) > \psi(\sum_{i=1}^N  [n_{xz}(i)+1]).
		\end{equation}
	According to the definition of digamma function, we can rewrite the above Equation as
		\begin{equation}
		\frac{1}{N} \sum_{i=1}^N \psi [n_{xy}(i)+1] > \frac{1}{N} \sum_{i=1}^N \psi [n_{xz}(i)+1].
		\end{equation}	
		
	\noindent Moreover, since the average distances of the points in $\{x_n, y_{n-\tau_1}\}$  to their $K$-nearest neighbor is smaller than those of the points in $\{x_n, z_{n-\tau_2}\}$ to their $K$-nearest neighbor, 
	
				\begin{equation}\label{AvgN2}
				\sum_{i=1}^N  [n_z(i)+1] > \sum_{i=1}^N  [n_y(i)+1].
				\end{equation}
		
\noindent Using the same analogy as above, we can conclude that
		\begin{equation}
		\frac{1}{N} \sum_{i=1}^N \psi [n_z(i)+1] > \frac{1}{N} \sum_{i=1}^N \psi [n_y(i)+1].
		\end{equation}	
		
\noindent Subsequently, based on the estimation of mutual information in Equations (\ref{EsMI1})-(\ref{EsMI2}), we can conclude that $I(x_n;y_{n-\tau_1})>I(x_n;z_{n-\tau_2})$. This implies that the causal interaction between two time series $X=\{x_n \}_{n=1}^N$ and  $Y=\{y_n \}_{n=1}^N$ with delay $\tau_1$ is higher than the causal interactions between $X=\{x_n \}_{n=1}^N$ and  $Z=\{z_n \}_{n=1}^N$ with delay $\tau_2$.

\end{proof}


\section*{Acknowledgment}

The authors would like to thank Dr Tomaso Poggio and Dr Leyla Isik from McGovern Institute for Brain Research at MIT, for kindly providing a dataset of MEG recordings for this study.

\ifCLASSOPTIONcaptionsoff
  \newpage
\fi



%
%
%
\bibliographystyle{ieeetr}
\bibliography{citations2016}

\end{document}